\documentclass{article}
\usepackage{spconf,amsmath,epsfig}
\usepackage{url}
\usepackage[hidelinks]{hyperref}
\usepackage{multirow}
\usepackage{caption}
\usepackage{tikz}

\usepackage{tablefootnote}
\usepackage[utf8]{inputenc}
\usepackage{array}
\usepackage{placeins}
\usepackage{afterpage}
\usepackage{subfig}
\usepackage{graphicx}
\usepackage{subfig}
\usepackage{amsmath,amsfonts,amssymb,amsthm}
\usepackage{mathtools}
\usepackage{csvsimple}
\usepackage{changepage}
\newcommand\MyBox[2]{
	\fbox{\lower0.75cm
		\vbox to 1.7cm{\vfil
			\hbox to 1.7cm{\hfil\parbox{1.4cm}{#1\\#2}\hfil}
			\vfil}%
	}%
}

\usepackage{tabularx}
    \newcolumntype{L}{>{\raggedright\arraybackslash}X}
\newcolumntype{M}{>{\centering\arraybackslash}m{5.5cm}}
\newcolumntype{N}{>{\centering\arraybackslash}m{1.5cm}}

\title{Application of Generative Adversarial Network (GAN) for Synthetic Training Data Creation to improve performance of ANN Classifier for extracting Built-Up pixels from Landsat Satellite Imagery}
\name{Amritendu Mukherjee, Dipanwita Sinha Mukherjee, Parthasarathy Ramachandran}
\address{Indian Institute of Science, Bangalore}
\begin{document}
%
\maketitle
\begin{abstract}
 Training a neural network for pixel based classification task using low resolution Landsat images is difficult as the size of the training data is usually small due to less number of available pixels that represent a single class without any mixing with other classes. Due to this scarcity of training data, neural network may not be able to attain expected level of accuracy. This limitation could be overcome using a generative network that aims to generate synthetic data having the same distribution as the sample data with which it is trained. In this work, we have proposed a methodology for improving the performance of ANN classifier to identify built-up pixels in the Landsat$7$ image with the help of developing a simple GAN architecture that could generate synthetic training pixels when trained using original set of sample built-up pixels. To ensure that the marginal and joint distributions of all the bands corresponding to the generated and original set of pixels are indistinguishable, non-parametric Kolmogorov–Smirnov Test and Ball Divergence based Equality of Distributions Test have been performed respectively. It has been observed that the overall accuracy and kappa coefficient of the ANN model for built-up classification have continuously improved from $0.9331$ to $0.9983$ and $0.8277$ to $0.9958$ respectively, with the inclusion of generated sets of built-up pixels to the original one.
\end{abstract}
\vspace{-0.5cm}
\section{Introduction}
\label{intro}
\vspace{-0.25cm}
In recent times, there has been a lot of research activities in the area of Generative Models\cite{oussidi2018deep,ruthotto2021introduction} as it demonstrates great potentials to improve existing state of the technology in various areas like Language Models, Image \& Video Generation, Code Generation, Speech Processing etc. The fundamental problems that these models try to solve are, (i) to estimate the underlying distribution of the observed data and (ii) to generate sample data from the same. The evolutionary journey of generative models encompasses the Gaussian Mixture Models\cite{viroli2019deep} (GMM), Auto Encoders, Variational Auto Encoders\cite{kingma2013auto} (VAEs), Generative Adversarial Networks\cite{goodfellow2020generative} (GANs) and Denoising Diffusion Probabilistic Models\cite{ho2020denoising,luo2022understanding} (DDPMs). By generating synthetic data from the estimated distribution, these generative models can help to improve the limitations of existing Machine Learning (ML) models that do not have enough training Data required to achieve desired level of accuracy.\\
Pixel-based Land Use Land Cover (LULC) classification from Landsat\footnote{\url{https://landsat.gsfc.nasa.gov/data/}; accessed on $15$ January, $2025$} data (with resolution of $30$m$\times30$m, i.e. each pixel represents an area of $30$m$\times30$m) using Artificial Neural Network (ANN), has been challenging as it requires pure quality of training Data where each pixel belongs to a particular class only. Otherwise, presence of pixels that represent a mixture of multiple classes in the training data, might downgrade the classification accuracy significantly. However, due to the poor resolution, it is very difficult to get sufficient training data with desired quality and thus, it is extremely hard to achieve high level of accuracy for the ANN model developed using the same. In order to overcome this problem, we've proposed a simple GAN architecture to generate synthetic training Data using good quality training pixels and have used the same to improve the classification accuracy of the ANN model for differentiating Built-Up and Non Built-Up regions in the Landsat$7$ satellite imagery. 
\vspace{-0.25cm}
\section{Data \& Study Area}
\label{data_study_area}
\begin{figure}[!h]
	\vspace{-0.5cm}
	\captionsetup{justification=centering,singlelinecheck=false,margin=1cm,format=hang}
	\subfloat{\label{truecolor_jaipur}\includegraphics[width=1.5in]{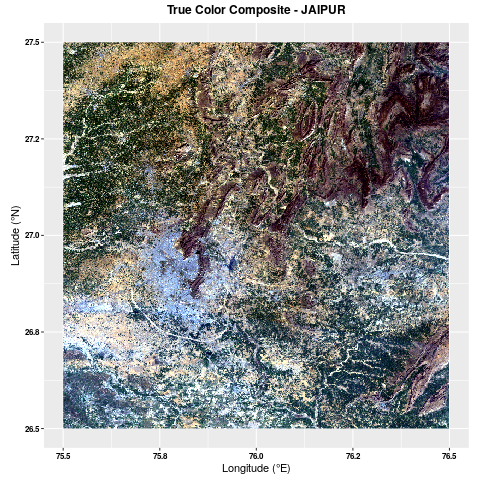}
	}
	\hspace*{1.5em}
	\subfloat{\label{falsecolor_jaipur}\includegraphics[width=1.5in]{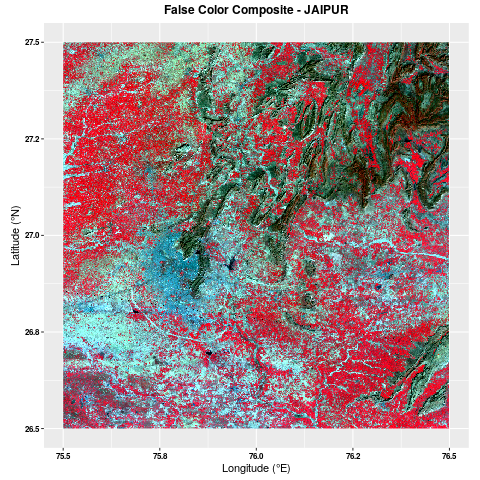}
	}
    \caption{True \& False Color Composite - Jaipur (Landsat7 image : 31-01-2017)}
    \label{true_false_color_composite_jaipur}
\end{figure}

\vspace{-0.25cm}
In order to investigate the classification performance of the proposed method, Landsat$7$ Enhanced Thematic Mapper Plus (ETM+) image corresponding to January $2017$ is sourced from U.S.G.S Earth Explorer\footnote{\url{https://earthexplorer.usgs.gov/}; accessed on $15$ January, $2025$}. The area of the study site is approximately $12100$ km$^2$($1^\circ$ Latitude$\times1^\circ$ Longitude) and contains Jaipur which is the capital and the largest city of the state of Rajasthan in the northern part of India. It's latitude and longitude span from $26.5^\circ$N to $27.5^\circ$N and $75.5^\circ$E to $76.5^\circ$E respectively. Estimated\cite{Rose2018} population density ((/$30''\times30''\approx1$km$^2$)) and total population corresponding to the year $2017$ are $530.23$ and $7.635$ millions respectively for the considered study site.
True and False Color Composite images of the study site have been provided in Figure~\ref{true_false_color_composite_jaipur}
\vspace{-0.25cm}
Google Earth Engine\footnote{\url{https://earthengine.google.com/}; accessed on $15$ January, $2025$}(GEE) for the same time period has been used to create training and testing set of manually verified sample pixels from the study site for both Built-Up and Non Built-Up classes. Training set comprises of a total number of $100$ Built-Up pixels and $400$ Non Built-Up pixels ($200$ Green\footnote{consists of equal representations from Forest and Vegetation classes}, $100$ Barren Land and $100$ Water). Testing set comprises a total number of $2000$ Built-Up pixels and $5000$ Non Built-Up pixels ($2000$ Green, $2000$ Barren Land and $1000$ Water)
\vspace{-0.5cm}
\section{Methodology}
\label{methodology}
\begin{figure}
	\hspace{-1.6cm}
	\captionsetup{justification=centering,singlelinecheck=false,margin=1cm,format=hang}
	\includegraphics[width=4.25in]{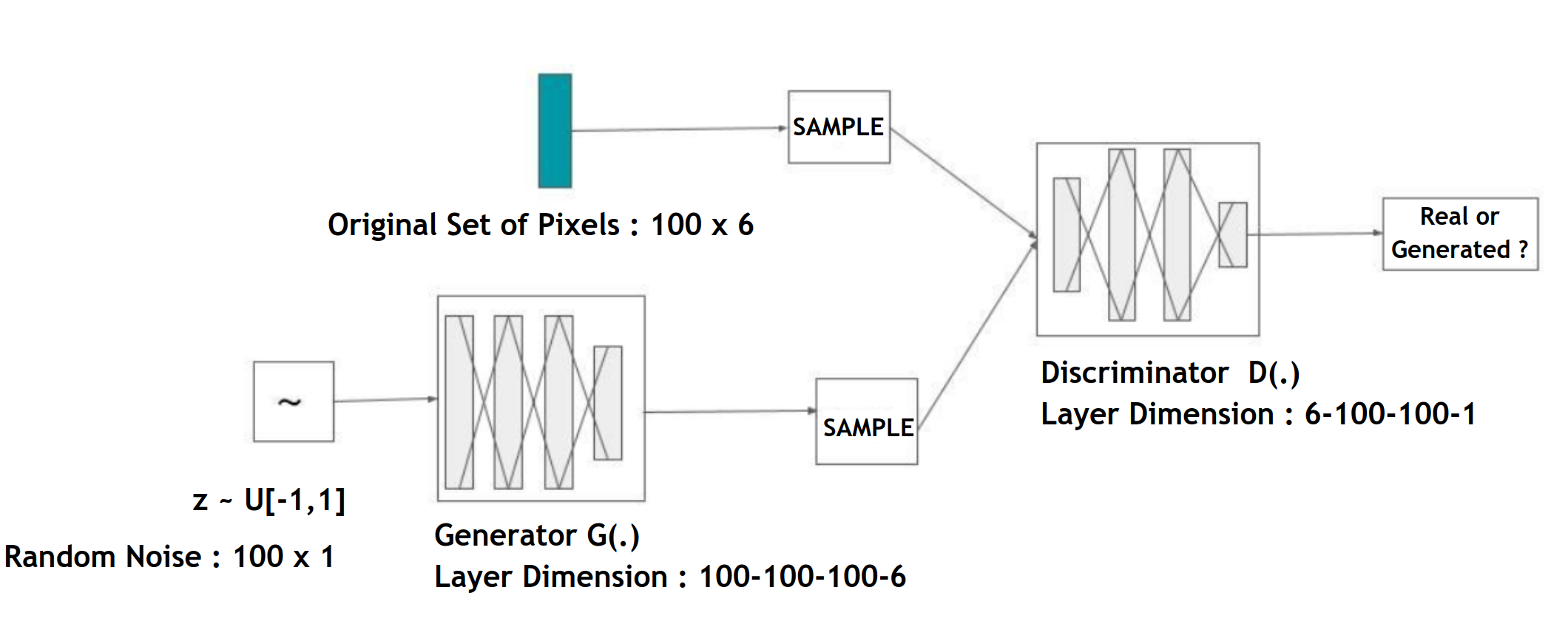}
	
    \caption{Architecture of the Implemented GAN}
    \label{gan_implementation}
\end{figure}

First, we've developed a simple GAN architecture to generate synthetic Built-Up pixels using actual Built-Up pixels from the original training set as described in Section~\ref{data_study_area}. In the GAN framework, for the purpose of learning the distribution $p_{g}$, a differentiable function $G(z;\theta_{g})$ represented by a neural network is crafted to map the input noise variables $p_{z}(z)$ to the data ($X$) space. Also, another neural network $D(x;\theta_{d})$ is developed to learn the probability of $x\in X$ coming from the actual data $X_{real}$ (and not from $p_{g}$). Where $D(.)$ is trained to maximize the probability of assigning correct labels to both the training data ($\in X_{real}$) as well as the generated ones (sample from the output of $G(z)$), $G(.)$ is trained to minimize the loss function $log(1-D(G(z))$ so that it learns to fail the discriminator $D(.)$ to assign the correct labels to the generated data (i.e. marks generated data as the real ones). Therefore, overall objective of GAN could be defined as the minimax game (equation~\ref{gan_eqn}) for the value function $V(G,D)$
\begin{equation}
\hspace{-0.25cm}
\footnotesize{\min_{G}\max_{D}V(G,D)=\mathbb{E}_{x\sim p_{data}(x)}[logD(x)]+\mathbb{E}_{z\sim p_{z}(z)}[log(1-D(G(z)]}
\label{gan_eqn}
\end{equation}
After experimenting with many architectures, in the GAN model that we have finally developed in this study for the purpose of generating synthetic Built-Up pixel using original set of $100$ Built-Up pixels, we've used band information (B1-BLUE, B2-GREEN, B3-RED, B4-NIR, B5-SWIR1, B6-SWIR2) of each original Built-Up pixel of the Landsat$7$ satellite image as input data ($X_{real}$) to be provided to the discriminator $D(.)$. The random noise ($z$) variable that has been given to the generator $G(.)$, is sampled from a Uniform distribution, i.e. $z\sim U[-1,1]$. Both $G(.)$ and $D(.)$ are designed as neural networks with $4$ layers including the input ones. For the generator and the discriminator, number of nodes (from input/1st Layer to the final output/4th layer) are $100-100-100-6$ and $6-100-100-1$ respectively. For both networks ($G(.)$ and $D(.)$), the activation functions are same, i.e. Sigmoid, ReLU and ReLU in the same order from input to output layers. A schematic diagram of the deployed GAN architecture along with dimensions of input noise and original data, has been demonstrated in Figure~\ref{gan_implementation}.\\
With the help of this GAN model, once trained, we have generated $3$ sets with each set having $100$ synthetic Built-Up pixels from the original training set of $100$ Built-Up pixels. In order to ensure that each set of generated Built-Up pixels and the original ones represent the same distribution, we've performed nonparametric Kolmogorov–Smirnov test\cite{massey1951kolmogorov} for each bands (e.g. distribution of the B1 band of the original data is compared with the same corresponding to the each set of generated ones). In addition, the joint distributions of all the $6$ bands (B1-B6) together corresponding to the original and each set of generated synthetic data, have also been compared using non-parametric Ball Divergence test\cite{pan2018ball} for equality of multivariate distributions.\\
Next, we have developed a simple ANN classifier to separate Built-Up pixels from the Non Built-Up ones. The ANN classifier is initially trained using the original $100$ Built-Up pixels and $400$ Non Built-Up pixels from the training set. Thereafter, each set of generated Built-Up pixels are gradually added to the training set of Built-Up pixels to train the ANN model. It could be noted here that the original set of Non Built-Up pixels has been kept constant throughout the experiment. Each time (starting from the original training set to the final one that consists of $100$ original Built-Up pixels along with $300$ generated ones), the performance of the ANN model is observed for the test data (comprises of $2000$ Built-Up pixels and $5000$ Non Built-Up pixels) in order to understand the impact of addition of generated Built-Up pixels to the training data.\\
In order to obtain the right set of parameters for achieving optimal accuracy of the ANN model for each time of training, grid search has been performed along with $10$ fold cross validation. It could be noted here that only one hidden layer has been used in all scenarios as the universal approximation theorem\cite{hornik1989multilayer} ensures that any bounded continuous function could be approximated arbitrarily well by a neural network with at least $1$ hidden layer having finite number of weights. We have noted that for each time of training the ANN classifier, grid search method has finalized the number of units in the hidden layer as $2$. However, from one configuration to another, weight decay\cite{goodfellow2016deep} ($\lambda$) parameter varies slightly (from $0.1$ to $0.4$). Weight decay, is primarily a $L_{2}$ regularization method
where the penalty on the $L_{2}$ norm of the weights is added to the original loss function (to discourage high values for weights) as shown in equation~\ref{eqn_weight_decay} where $L(.)$ is the loss function, $w$ is the weight matrix and $\lambda$ is the parameter that defines the strength of the penalty.
\begin{equation}
\footnotesize{L_{updated}(w) = L_{original}(w)+\lambda w^\intercal w}
\label{eqn_weight_decay}
\end{equation}
Performance measures of the ANN model are derived from confusion matrix\cite{Ting2017}, which are two dimensional matrices with one dimension representing the true class of the pixel and the other dimension representing the predicted class by the classifier. Set of test pixels is used to calculate the accuracy measures which are sensitivity, specificity, positive prediction value (PPV), negative prediction value (NPV) and total accuracy. In addition to these accuracy measures, Cohen's Kappa coefficient\cite{cohen1960coefficient} ($\kappa$) is also reported to show the conformance of classified results with the ground truth. $\kappa$ coefficient checks whether the results of the ANN classifier is in agreement with the ground truth represented by testing set of pixels for both Built-Up and Non Built-Up classes.\\
All computations have been performed with the help of R\footnote{\url{https://www.r-project.org/}; accessed on $15$ January, $2025$} and Python\footnote{\url{https://www.python.org//}; accessed on $15$ January, $2025$} software along with associated frameworks and packages like TensorFlow (Python), NumPy (Python), pandas (Python), Matplotlib (Python), Caret (R), Ball (R) etc. as required.
\vspace{-0.5cm}
\section{Results \& Discussions}
\label{results_discussions}
\begin{figure}[!h]
	\vspace{-0.5cm}
	\hspace{-1.5cm}
	\captionsetup{justification=centering,singlelinecheck=false,margin=1cm,format=hang}
	\subfloat[With only original (100) training pixels] {\label{truecolor_jaipur}\includegraphics[width=2.25in]{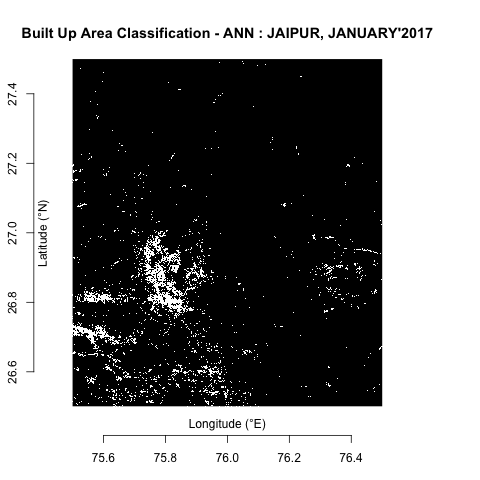}
	}
	\hspace*{-0.5cm}
	\subfloat[With original (100) and \\ \hspace{-0.5cm} generated (300) training pixels]{\label{falsecolor_jaipur}\includegraphics[width=2.25in]{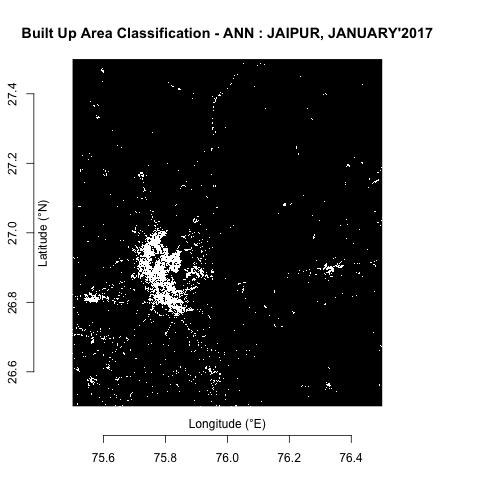}
	}
    \caption{Built-Up Classification using ANN \\ \hspace{-1.5cm} (with and without addition of generated pixels)}
    \label{ann_classifier_1_2}
\end{figure}

As discussed previously in Section~\ref{methodology}, p-values associated with the Kolmogorov–Smirnov (KS) test that has been performed to confirm that the individual bands corresponding to each set of generated Built-Up pixels and the same bands from the original set of Built-Up pixels represent same distribution, have been reported in Table~\ref{ks_test}.
\begin{table}[!t]
    \captionsetup{justification=centering}
	\caption{{\\ \hspace{1.0cm}\fontsize{8}{10}\selectfont{\MakeUppercase{\\ $p$ Value of Kolmogorov–Smirnov Test \\ Individual Bands - Generated \& Original Set}}}} 
    \centering 
	\renewcommand{\arraystretch}{1.5}
	\footnotesize
    \scriptsize
    \begin{tabular}{|c|c|c|c|} 
		
        \hline
		\multirow{2}{*} {\shortstack{\textbf{Landsat $7$ Bands}}} & \multicolumn{3}{c|}{\textbf{Compared Set of Pixels}} \\ \cline{2-4} 
		& \textbf{\shortstack{\\Generated Set $1$ \& \\
Original Set}}&  \textbf{\shortstack{\\Generated Set $2$ \& \\
Original Set}} &  \textbf{\shortstack{\\Generated Set $3$ \& \\
Original Set}} \\
		\hline
        \hline
        B1 (BLUE) & $0.5806$ & $0.2106$ & $0.4676$ \\
        \hline
        B2 (GREEN) & $0.4676$ & $0.6994$ & $0.5806$ \\
        \hline
        B3 (RED) & $0.9062$ & $0.4676$ & $0.8127$ \\
        \hline
        B4 (INR) & $0.4676$ & $0.8127$ & $0.1545$ \\
        \hline
        B5 (SWIR1) & $0.4676$ & $0.8127$ & $0.5806$ \\
        \hline
        B1 (BLUE) & $0.8127$ & $0.9062$ & $0.8127$ \\
        \hline
	
		\hline
 
	\end{tabular}
	\label{ks_test}
\end{table}	
As the noted $p$ values (for all the observations in Table~\ref{ks_test}) are significantly high ($>0.05$), null hypothesis of KS test (H$_{0}$ : Two samples compared come from the same continuous distribution) can't be rejected and therefore, it could be concluded that marginal distributions of individual bands for each set of pixels generated with the help of GAN architecture (Figure~\ref{gan_implementation}) implemented, are same as that of corresponding bands of the original set.\\
Similarly, p-values associated with the non parametric Ball Divergence test that has been performed to confirm that the multivariate joint distribution of all $6$ bands corresponding to each set of generated pixels and the joint distribution of all bands from the original set are same, have been reported in Table~\ref{ball_divergence}. As the $p$ values reported in Table ~\ref{ball_divergence} are high ($>0.05$), the null hypothesis (H$_{0}$ : Two distributions of samples compared are not distinct) could not be rejected and thus, we could infer that the joint distributions of all the bands together for each set of generated pixels, are same as that of the original one.
\hspace{-1.5cm}
\begin{table}[!t]
	\captionsetup{justification=centering}
	\caption{{\\ \hspace{1.0cm}\fontsize{8}{10}\selectfont{\MakeUppercase{\\ Comparison of Joint Distribution of Landsat 7 Bands \\ Ball Divergence Test : Generated \& Original Set}}}} 
    \centering 
	\renewcommand{\arraystretch}{1.5}
	\footnotesize
    \scriptsize
    \begin{tabular}{|c|c|} 
		\hline
		\textbf{Compared Sets} & \textbf{$p$ Value}\\ 
        \hline
        \hline
        Generated Set $1$ \& Original Set & $0.35$ \\
        \hline
        Generated Set $2$ \& Original Set & $0.54$ \\
        \hline
        Generated Set $3$ \& Original Set & $0.23$ \\
        \hline

	\end{tabular}
	\label{ball_divergence}
\end{table}\\
Next, we've incrementally added these sets of generated Built-Up pixels (having identical marginal and joint distributions with the original set of pixels for all bands) to the original training set and have noted the performances in terms of Sensitivity, Specificity, PPV, NPV, Accuracy and Kappa ($\kappa$) Coefficient for the testing set as reported in Table~\ref{ann_performance} which also include final parameters used in the ANN model.
\begin{table}[!t]
\begin{adjustwidth}{-1.0cm}{-1.0cm}
	\caption{{\fontsize{8}{10}\selectfont{\MakeUppercase{\\ Accuracy Measures : ANN Classifier}}}} 
    \centering 
	\renewcommand{\arraystretch}{1.5}
    \hskip-1.0cm
    \resizebox{0.6\textwidth}{!}{
\begin{tabular}{|M|c|c|N|N|c|N|}\hline
\textbf{Training Configuration} & \textbf{Sensitivity} & \textbf{Specificity} & \textbf{PPV} & \textbf{NPV} & \textbf{Accuracy} & \textbf{Kappa($\kappa$)} \\\hline \hline
Built-Up Pixels (Original) : 100 \hspace{2.0cm} Non Built-Up Pixels (Original) : 400 \hspace{2.0cm} No. of Nodes (ANN) : 2 \hspace{2.0cm} Weight Decay $\lambda$ (ANN) : $0.4$ & $0.9860$ & $0.8010$ & $0.9253$ & $0.9581$ & $0.9331$ & $0.8277$ \\\hline
Built-Up Pixels (Original) : 100 \hspace{2.0cm} Built-Up Pixels (Generated) : 100 \hspace{2.0cm} Non Built-Up Pixels (Original) : 400 \hspace{2.0cm} No. of Nodes (ANN) : 2 \hspace{2.0cm} Weight Decay $\lambda$ (ANN) : $0.3$ & $0.9852$ & $0.8795$ & $0.9534$ & $0.9596$ & $0.9550$ & $0.8869$ \\\hline
Built-Up Pixels (Original) : 100 \hspace{2.0cm} Built-Up Pixels (Generated) : 200 \hspace{2.0cm} Non Built-Up Pixels (Original) : 400 \hspace{2.0cm} No. of Nodes (ANN) : 2 \hspace{2.0cm} Weight Decay $\lambda$ (ANN) : $0.1$ & $0.9906$ & $0.9280$ & $0.9717$ & $0.9753$ & $0.9727$ & $0.9322$ \\\hline
Built-Up Pixels (Original) : 100 \hspace{2.0cm} Built-Up Pixels (Generated) : 300 \hspace{2.0cm} Non Built-Up Pixels (Original) : 400 \hspace{2.0cm} No. of Nodes (ANN) : 2 \hspace{2.0cm} Weight Decay $\lambda$ (ANN) : $0.1$ & $0.9994$ & $0.9955$ & $0.9982$ & $0.9985$ & $0.9983$ & $0.9958$ \\\hline
 \end{tabular}
 }
	\label{ann_performance}
\end{adjustwidth}
\end{table}	
It could easily be noted in Table~\ref{ann_performance} that for all the accuracy measures considered, the performance of the ANN model has steadily improves with the inclusion of generated set of built-up pixels to the original one. To elaborate, the corresponding values of accuracy and $\kappa$ have increased from from $0.9331$ to $0.9983$ and $0.8277$ to $0.9958$ respectively and the same observation is applicable for other accuracy measures (e.g. Sensitivity, Specificity, PPV \& NPV) as well.\\ 
Additionally, we also have performed a visual comparison between the true/false color  composite images (Figures~\ref{true_false_color_composite_jaipur}) of the entire study site and the classified built-up obtained using ANN classifier (with only original training pixels and with the addition of generated pixels to the original ones), as shown in  Figure~\ref{ann_classifier_1_2}. The classified built-up images, as shown in Figure~\ref{ann_classifier_1_2} reveal that the ANN classifier with the addition of generated Built-Up pixels to the original ones performs significantly better with less misclassification compared to the ANN classifier trained using only the original set of Built-Up pixels.
\vspace{-0.25cm}
In this work, we've developed a simple generative model and have explored it's application to generate synthetic training data where the available training data is less and thereby, using the same, could help to train a neural network efficiently to achieve desired level of accuracy. It could be noted here that for the purpose of demonstrating the concept, we have developed a simple GAN architecture but any other generative models (like VAE, Diffusion models etc.) could also be developed for the same purpose. However, training a large generative network for generating samples from a small set of low dimensional training dataset, might be unnecessary and computationally expensive. Similarly, to illustrate the proposed methodology, multi-spectral Landsat$7$ data has been used but the same idea could be applied to other types of satellite images as well.
\vspace{-0.25cm}
\bibliographystyle{IEEEbib}
\bibliography{Ref.bib}
\end{document}